\documentclass{article}
\usepackage{times}
\usepackage{url}
\usepackage{amsmath}

\title{Computational Argumentation and Cognition\thanks{We wish to thank all the speakers and the participants of the COGNITAR 2020 workshop for taking part in the discussions at the workshop and for commenting on earlier drafts of this paper.}}

\author{Emmanuelle Dietz$^\text{1}$, 
Antonis Kakas$^\text{2}$, Loizos Michael$^\text{3}
$\bigskip \\
\small $^\text{1}$ International Center for Computational Logic, TU
 Dresden, Germany \\ \small emmanuelle.dietz@tu-dresden.de \\
\small $^\text{2}$ 
 Department of Computer Science, University of Cyprus, Nicosia, Cyprus \\ \small{antonis@ucy.ac.cy}\\
\small $^\text{3}$ 
Open University of Cyprus \& CYENS Center of Excellence, Cyprus \\ \small {loizos@ouc.ac.cy}
}

\date{}

\begin{document}

\maketitle

\begin{abstract}
This paper examines the interdisciplinary research question of how to integrate Computational Argumentation, as studied in AI, with Cognition, as can be found in Cognitive Science, Linguistics, and Philosophy. It stems from the work of the 1st Workshop on Computational Argumentation and Cognition (COGNITAR), which was organized as part of the 24th European Conference on Artificial Intelligence (ECAI), and took place virtually on September 8th, 2020.
The paper begins with a brief presentation of the scientific motivation for the integration of Computational Argumentation and Cognition, arguing that within the context of Human-Centric AI the use of theory and methods from Computational Argumentation for the study of Cognition can be a promising avenue to pursue.
A short summary of each of the workshop presentations is given showing the wide spectrum of problems where the synthesis of the theory and methods of Computational Argumentation with other approaches that study Cognition can be applied.
The paper presents the main problems and challenges in the area that would need to be addressed, both at the scientific level but also at the epistemological level, particularly in relation to the synthesis of ideas and approaches from the various disciplines involved.
\end{abstract}


\section*{Introduction}

Artificial Intelligence today requires the synthesis of ideas and methods from a wide variety of disciplines. This need stems primarily from the expectation that AI systems will have a natural behavior at the cognitive level of their human users, and will exhibit human-like cognitive abilities, both at the level of their internal operation and at the level of interacting with humans, or between themselves. Understanding and automating Cognition is, thus, of paramount importance for delivering successful AI systems.

COGNITAR 2020 (\url{http://cognition.ouc.ac.cy/cognitar/}) was the first event in a new series of International Workshops whose purpose is to help investigate the synthesis between Computational Argumentation in AI, with the study of Cognition in other disciplines such as Cognitive Science, Linguistics, and Philosophy. The aim is to promote and allow this integration to inform the development of AI systems, so that these systems are more attuned to a natural human-like behavior. The main general questions that concerned the workshop were the following:

\begin{itemize}
\item  Can argumentation provide the basis for computational models of human reasoning that are cognitively adequate?
\item How can we form a synthesis between computational argumentation and theories of cognition that will give us models of computational cognition for the development of AI systems?
\end{itemize}

Cognitively-inspired AI systems have recently attracted a renewed attention from both academia and industry, and the awareness about the need for additional research in this interdisciplinary field is gaining widespread acceptance. In particular, Human-Centric AI requires systems that are built based on a holistic understanding of the human mind and its various cognitive faculties at many different levels.
Cognition in AI systems will be important for several aspects of their design:

\begin{itemize}
\item Comprehending the data or information from their environment.
\item Deciding on what actions to take or recommend to users.
\item Learning new knowledge pertaining to their problem domain.
\item Explainability at different cognitive levels of expertise.
\item Social interaction with human users and artificial agents.
\item Ensuring an ethical operation and general behavior.
\end{itemize}

Understanding and automating the higher-level cognition involved in carrying out the above general tasks of human behavior requires a (logical) formalism that is close to human reasoning as carried out in everyday tasks. The formalism would need a high degree of flexibility to allow dynamic changes and the revision of drawn conclusions. A candidate such framework is that of (dialectic) argumentation as advocated in Philosophy and Cognitive Psychology in early work \cite{Perelman1969,Pollock1987,Pollock1995,Toumlin1958,Walton1996} and newly supported by many recent studies connecting argumentation with a wide spectrum of different aspects of human reasoning \cite{Bex2011,Blair2012,PolishManifesto2014,van2019handbook,Godden2015Rationality,Hinton2019LanguageAA,Hoffmann2016,Krabbe2017,MERCIER2016689,mercier:sperber:2011,Oaksford2020,OswaldEtal2018,Eemeren_grootendorst_2003,VisserKDKBR20,WaltonGordon2015,WaltonGordon2019,Zenker2018}, including such studies within the context of AI\footnote{See also references at the following website: \url{http://cognition.ouc.ac.cy/argument/}} \cite{Atkinson_EtAl_2017,baroni2018handbook,BudzynskaR19,Cramer19,DAVILAGARCEZ2014109,HornikxHahn2012,Informalizing2019,KakasMichael16,LietoV14,CramerEmma20,Michael16,Michael19,Prakken2011,ReedGrasso2007,ArielKraus2016,CA-Syll2019,Verheij16}. The main premise is that argumentation is native to human reasoning and can provide a foundation for (higher-level) cognition and, in turn, for human-centric AI. Can computational argumentation then form an underlying theoretical and practical basis for modeling cognition and building human-centric AI systems?

\section*{COGNITAR 2020}

In this section we briefly present the talks at the COGNITAR 2020 workshop aiming to convey, through these summaries, the specific research problem that each talk exposed within the scope of synthesizing Computational Argumentation and Cognition.

The workshop had two invited speakers\footnote{The first part of COGNITAR comprising the two invited talks was held jointly with the ECAI workshop ``Artificial and Human Intelligence: Formal and Cognitive Foundations for Human-Centred Computing''.}: Martin Hinton from the area of Argumentation and Natural Language and Jakob Suchan from the area of engineering cognitive solutions to Autonomous Systems. The workshop also included eight other talks and an open discussion on ``Argumentation and Human-Centric AI'' at the end of the meeting.

\paragraph*{Martin Hinton, \textit{``Order out of Chaos: A Systematic Approach to the Evaluation of Natural Language Argumentation''}}

Human cognition is best understood through language. Martin Hinton argued that systematic approaches to argumentation are of direct relevance to computational application and proposes the Comprehensive Assessment Procedure for Natural Argumentation, CAPNA. CAPNA, a systematic argument evaluation procedure, contains various stages, such as argument identification, informal argument pragmatics, and the underlying reasoning analysis. CAPNA then accepts an argument, if it has not been rejected at any of these stages. For the application of AI, it seems more important to understand the way arguments work and how they are expressed than the way people think.

\paragraph*{Jakob Suchan, \textit{``Commonsense Reasoning for Autonomous Driving''}}

Jakob Suchan presented a framework for visual sense making within the realm of autonomous driving. The challenge is to extract from the low level visual data high-level cognitive information about the observed state of the world. Thus the work addresses in a concrete way the task of integrating low-level with high-level processes that would result in a cognitive comprehension of the visual data, on which then other actions can be based. The extraction of high-level cognitive information is carried out through abductive explanations of the perceived scene based on common sense reasoning about the dynamics of the scenes. This uses an Event Calculus based representation and Answer Set Programming for the process of abduction. 

\paragraph*{Kenneth Skiba and Matthias Thimm, \textit{``Towards Ranking-based Semantics for Abstract Argumentation using Conditional Logic Semantics''}}
 
Conditional logics is a knowledge representation and reasoning formalism for capturing associations between properties of the world. Kenneth Skiba and Matthias Thimm investigated the integration of this formalism with abstract argumentation. They presented a technical solution for translating an abstract argumentation framework into conditional logics, and gave the semantics to the resulting translation by ranking the arguments in terms of how many possible worlds that satisfy the conditionals also satisfy each argument. Effectively, this understanding goes beyond the standard binary acceptability relation, but offers a ranking of the arguments in terms of their intuitive ``acceptability'' strength.

\paragraph*{Marcin Koszowy, Steve Oswald, Katarzyna Budzynska and Barbara Konat, \textit{``The Role of Rephrase in Argumentation: Computational and Cognitive Aspects''}}

Argumentation is influenced significantly by the environment in which it takes place. For example, simply asking to rephrase an argument or reformulating a speaker's own claim can have a non-trivial impact on the persuasive effectiveness within a dialogue. Marcin Koszowy et al.\ study this, aiming to both formulate a pragmatic theory for rephrase that combines linguistic evidence for rephrase use with examining empirically the effect of rephrasing. In particular, they have carried out an extensive corpus study combining expertise on linguistics and cognitive science which can be then applied in argument mining. They have presented their results on corpus analysis and of experiments that they have run with human participants to test explicitly the effect of rephrase on persuasion.

\paragraph*{Antonio Lieto and Fabiana Verner, \textit{``On the Impact of Fallacy-based Schemata and Framing Techniques in Persuasive Technologies''}}

Fallacies are powerful heuristics, which are incorrect but psychologically persuasive, because they appear 
psychologically plausible. Antonio Lieto and Fabiana Verner presented two studies where some well-known fallacious reducible schemata were applied in the area of technology-based persuasion.
The results of the first study within e-commerce indicate that half of the users were not influenced by the used fallacies, whereas the other half were influenced by some of them.
The second study was within an online magazine of a university, and showed that the Argumentum Ad Populum fallacy (a proposition must be true because many or most people believe it) was not efficient. 
They also tested the framing effect, a technique where the same content is formulated differently, and found out that the number of visualization and clicks were higher with negative framing. Finally, the application of the fallacious reducible schemata is helpful to analyze the behavior of different users and allows to measure which mix of techniques can be more efficient.

\paragraph*{Sangeet Khemlani, \textit{``Leveraging Cognitive Constraints for Interpretable AI''}}

Human thinking is extraordinary limited. It depends on mental simulations (mental models) which are very costly, constrained, and biased in various ways.
People use these models to reason about space, time, and causality.
Sangeet Khemlani presented the MReasoner, a unified computational implementation of the model theory, that simulates human reasoning errors and their optimal performance. One example demonstrates how reasoning between qualitative spatial relations within MReasoner interfaces with object recognition based on machine learning technologies.
Sangeet Khemlani argued that the insights gained by modeling the way that people reason will help design better AI and enrich interactions with AI and robotic agents.

\paragraph*{Katarzyna Budzynska and Marcin Koszowy, \textit{``An Empirical Turn in Studying Ethos in Argumentation''}}

Ethos, the character of the speaker, is one of the key elements of real-life argumentation in debates and more generally in human communication. Katarzyna Budzynska and Marcin Koszowy presented their initial work and the long-term program of a newly formed laboratory, “The New Ethos”, whose goal is to implement innovative technologies of ethos mining and ethos analytics. Their work is theoretically driven and empirically grounded in the analysis of large-scale corpora which contain real-life argumentation in different domains. They identify a variety of ethotic strategies which speakers use in natural communication with the goal of providing an insight into ethos dynamics.

\paragraph*{Marcos Cramer, \textit{``Comparing the Cognitive Plausibility of Abstract Argumentation Semantics Based on Empirical Studies''}}

Abstract argumentation frameworks are amenable to multiple semantics, each capturing and formalizing a different intuition of when a set of arguments is collectively considered to be acceptable. Naturally, one may wish to investigate which of those intuitions are cognitively plausible. Marcos Cramer discusses empirical studies that seek to identify which argumentation semantics predicts best how humans evaluate arguments. The studies point towards certain conclusions about which family of semantics is more predictive of human choices, but also suggest that further studies might be necessary to distinguish among the individual members of that family, or to investigate additional aspects of argumentation semantics (e.g., the existence of multiple acceptable sets for a given argumentation framework).

\paragraph*{Ilya Ashikhmin, \textit{``Debatt --- The Social Argument Map''}}

Ilya Ashikhmin presented Debatt, the social argument map project whose purpose is to develop a web application for constructing argument maps. 
The motivation comes from observations in social networks, where civilized discussions do not seem to be encouraged and good arguments/ counterarguments are often scattered across multiple unconnected discussions. Formal systems do not seem to be suitable as they are too complex or too restrictive, which is not necessary for the purpose. 
With Debatt, Ilya Ashikhmin aims at finding a good balance between keeping the web application simple while allowing for extensions and the flexibility to decide which formalism to apply for acceptability of arguments. 

\paragraph*{Annemarie Borg and Floris Bex, \textit{``Good Explanations for Formal Argumentation''}}

Explainability is today one of the most important requirements on (human-centric) AI systems. Explanations are closely related to argumentation and Annemarie Borg and Floris Bex presented some general desiderata for ``good'' explanations in formal argumentation frameworks. For example, such explanations would need to be compatible with the cognitive level of the intended users of applications build on formal argumentation. They also pointed out that the form and generation of argumentation-based explanations would benefit from the wealth of studies on explanation in social sciences.

\section*{Challenges and Future Directions}

In this section we present the main future challenges in the quest to bring closer Computational Argumentation and Cognition as these arose from the talks of the workshop and an open discussion on the links between Computational Argumentation and Human-Centric AI around the following two general questions:

\begin{itemize}

\item At the scientific level: \textit{``What is the potential role of Computational Argumentation for Human-Centric AI and how can this role be realized?''}

\item At the epistemological level: \textit{``What are the main synergies between disciplines needed to harness the potential of 
computational argumentation for Human-Centric AI?''}

\end{itemize}

The main purpose of the discussion was to raise the various challenges and problems that would need to be addressed to answer in a constructive way these questions. More specifically, the following issues were identified:

\begin{itemize}

\item Does the natural link of argumentation to human cognition make its use in building AI systems necessary? 

In particular, how can argumentation technology facilitate a meaningful engagement between AI and human users? How can we harness the link of argumentation with human reasoning in Natural Language for this purpose?

\item What is the role of argumentation, if any, in linking the System 1 and System 2 cognitive processes?

In particular, can argumentation be used to build on top of System 1 modules of AI systems, e.g., sub-symbolic connectionist approaches to machine learning, a System 2 cognitive model?

\item How can argumentation help with the ethical requirements of AI systems and indeed with the responsible use of such systems?

In particular, how can argumentation help regulate the tension between persuasiveness and ethical behavior of AI systems? Can argumentation technology mitigate the effect of non-responsible use of AI systems by exposing marketing practices based on fallacious arguments?

\item What is the role of argumentation in the quest for Explainable AI? 

In particular, in the context of Machine Learning, can argumentation provide a layer on top of learning that would allow the further use of the results of learning?

\item What empirical studies will help us understand the link between human reasoning and computational argumentation is a way useful for building AI systems?

In particular, how can we design such experiments where the reasons for and against considered by humans in their reasoning are revealed by them without any bias from the experimental setup?

\item What are the main synergies with other disciplines that would allow computational argumentation to better address the needs of Human-Centric AI?

In particular, how can we harness the results and methods from the various extensive studies of argumentation in other disciplines whose focus comes from the use of argumentation in human introspection and/or debate?

\end{itemize}

Admittedly, most if not all of the issues underlying the above questions constitute central concerns for Human-Centric AI in general. The purpose of this paper, and indeed that of the COGNITAR workshop series, is to study these concerns from the perspective of argumentation, and to examine how this perspective can contribute, alongside other approaches, in addressing the central challenges of Human-Centric AI.

\small 
\bibliographystyle{plain}
\bibliography{references}

\end{document}